\documentclass[journal]{IEEEtran}
\usepackage{url}
\usepackage{times}
\usepackage{epsfig}
\usepackage{graphicx}
\usepackage{amsmath}
\usepackage{amssymb}
\usepackage{algpseudocode}
\usepackage{algorithm}
\usepackage{algorithmicx}
\usepackage{multirow}
\usepackage{booktabs}
\usepackage{setspace}
\usepackage{float}
\usepackage{color}
\usepackage[square,sort,sort&compress,numbers]{natbib}
\ifCLASSINFOpdf
\else
\fi

\begin{document}

\title{A Uniform Representation Learning Method for OCT-based Fingerprint Presentation Attack Detection and Reconstruction}

\author{
Wentian Zhang, Haozhe Liu, Feng Liu$^*$, Raghavendra Ramachandra
\thanks{The correspondence author is Feng Liu and the email: feng.liu@szu.edu.cn.}
\thanks{Wentian Zhang, Haozhe Liu and Feng Liu are with 
the College of Computer Science and Software Engineering, Shenzhen University, Shenzhen 518060, China; SZU Branch, Shenzhen Institute of Artificial Intelligence and Robotics for Society, China, and also with Guangdong Key Laboratory of Intelligent Information Processing, Shenzhen University, Shenzhen 518060, China.}
\thanks{Raghavendra Ramachandra is with the Norwegian Biometrics Laboratory (NBL), Norwegian University of Science and Technology (NTNU), Gjøvik 2818, Norway.}
\thanks{This work was done when Haozhe Liu was a visiting student at NTNU, Norway.}
}

\maketitle
\begin{abstract}

The technology of optical coherence tomography (OCT) to fingerprint imaging opens up a new research potential for fingerprint recognition owing to its ability to capture depth information of the skin layers. Developing robust and high security Automated Fingerprint Recognition Systems (AFRSs) are possible if the depth information can be fully utilized. However, in existing studies, Presentation Attack Detection (PAD) and subsurface fingerprint reconstruction based on depth information are treated as two independent branches, resulting in high computation and complexity of AFRS building.Thus, this paper proposes a uniform representation model for OCT-based fingerprint PAD and subsurface fingerprint reconstruction. Firstly, we design a novel semantic segmentation network which only trained by real finger slices of OCT-based fingerprints to extract multiple subsurface structures from those slices (also known as B-scans). The latent codes derived from the network are directly used to effectively detect the PA since they contain abundant subsurface biological information, which is independent with PA materials and has strong robustness for unknown PAs. Meanwhile, the segmented subsurface structures are adopted to reconstruct multiple subsurface 2D fingerprints. Recognition can be easily achieved by using existing mature technologies based on traditional 2D fingerprints.
Extensive experiments are carried on our own established database, which is the largest public OCT-based fingerprint database with 2449 volumes. 
In PAD task, our method can improve 0.33\% Acc from the state-of-the-art method. 
For reconstruction performance, our method achieves the best performance with 0.834 mean Intersection of Union (mIOU) and 0.937 Pixel Accuracy (PA). By comparing  with the recognition performance on surface 2D fingerprints (e.g.  commercial and high resolution), the effectiveness of our proposed method on high quality subsurface fingerprint reconstruction is further proved.
\end{abstract}

\IEEEpeerreviewmaketitle

\section{Introduction}
\label{sec:Intro}

\begin{figure}[t]
  \centering
  \includegraphics[width=0.48\textwidth]{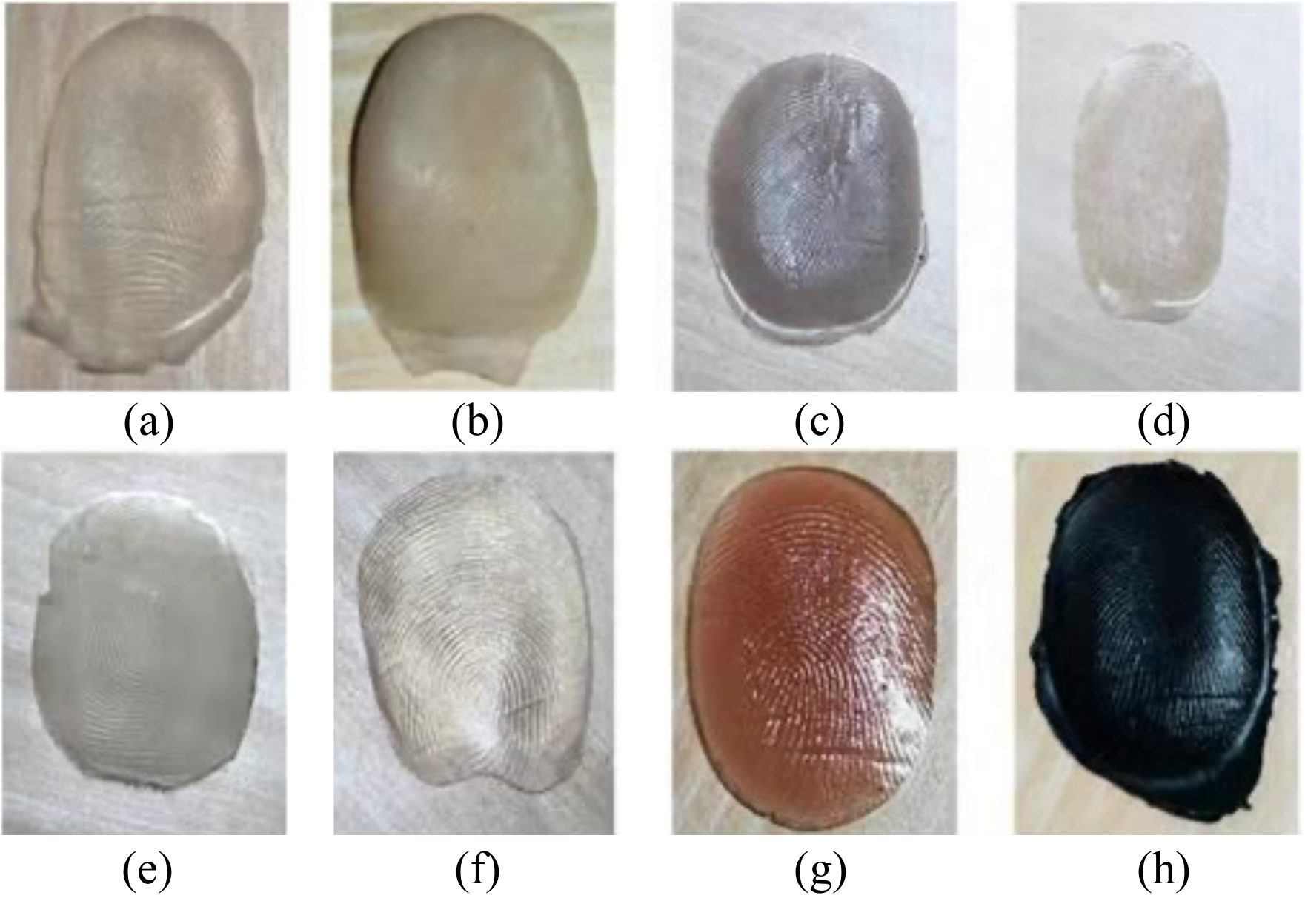}
  \caption{Some samples of the produced spoof fingers made from (a) silica gel. (b) silicone. (c) silicone with conductive silver. (d) ultrathin silicone. (e) glass adhesive. (f) PDMS. (g) epoxy resin. (h) capacitance glue.}
  \label{fig:pa}
\end{figure}

\IEEEPARstart{B}{iometrics} as a reliable authentication feature has been widely employed in the intelligent device terminals.
Among those biometric characteristics, fingerprint along with various Automated Fingerprint Recognition Systems (AFRSs) has been applied in forensics and security for centuries ~\cite{lin2018matching,das2018recent}. 
However, due to the nature of surface fingerprint imaging, there are two restrictions of traditional AFRSs in the real scenario.  
On the one hand, traditional AFRSs have low tolerance to poor-quality images, such as worn-out fingerprints and ultrawet/ dry fingers, which will badly affect the recognition accuracy. 
On the other hand, presentation attacks (PAs) are bringing a raising security problem to AFRSs, which caused concerns about the reliability of such systems~\cite{jia2007new,antonelli2006fake,lee2009fake,nikam2008fingerprint}. 
Even spoof fingers made from very low cost materials~\cite{liu2019high} can easily attack those AFRSs~\cite{goicoechea2016evaluation}. Fig.~\ref{fig:pa} shows some examples of spoof fingers with various materials and different manufacturing processes. We can find that the PAs are unpredictable due to their materials and craftsmanship~\cite{nogueira2016fingerprint,chugh2018fingerprint,chugh2019fingerprint,liu2021fingerprint,zhang2021face}. To build robust and high security AFRSs, it is essential to equip a PAD module.

\begin{figure*}[htpb]
  \centering
  \includegraphics[width=0.98\textwidth]{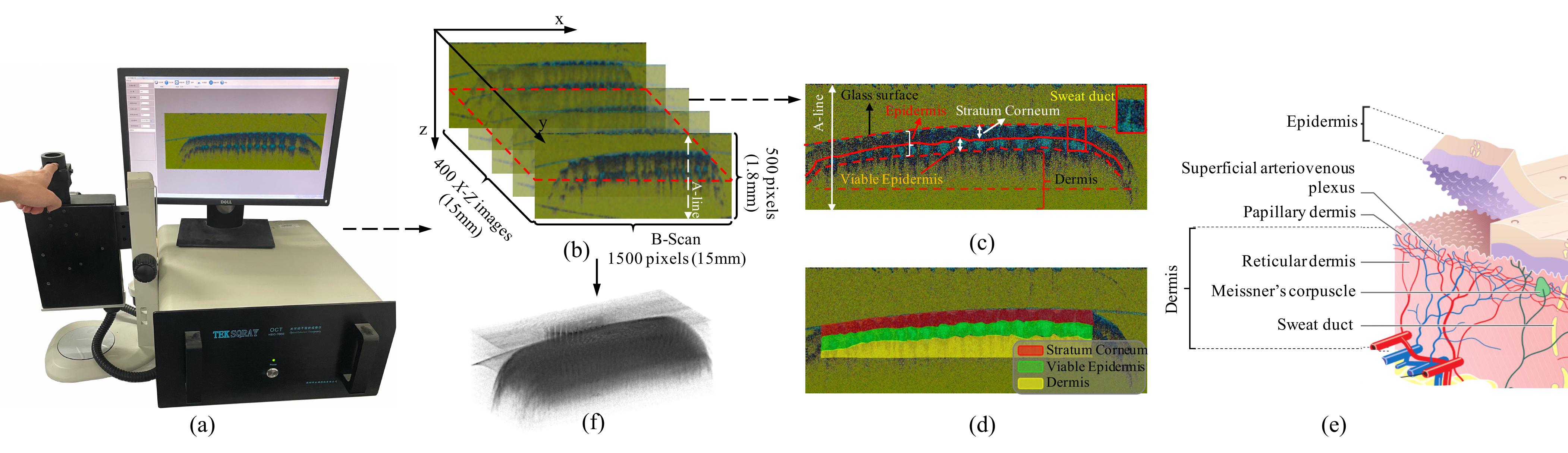}
  \caption{A sample of OCT-based fingerprint acquisition procedure. (a) OCT device. (b) Collected data. (c) A cross-section slice of OCT-based fingerprint (i.e. B-scan) . (d) B-scan split to three biological structure layers. (e) Skin biological structure diagram~\cite{skinlayer}. (f) 3D fingerprint reconstructed by (b) .}
  \label{fig:oct}
\end{figure*}
\begin{figure*}[h]
  \centering
  \includegraphics[width=0.99\textwidth]{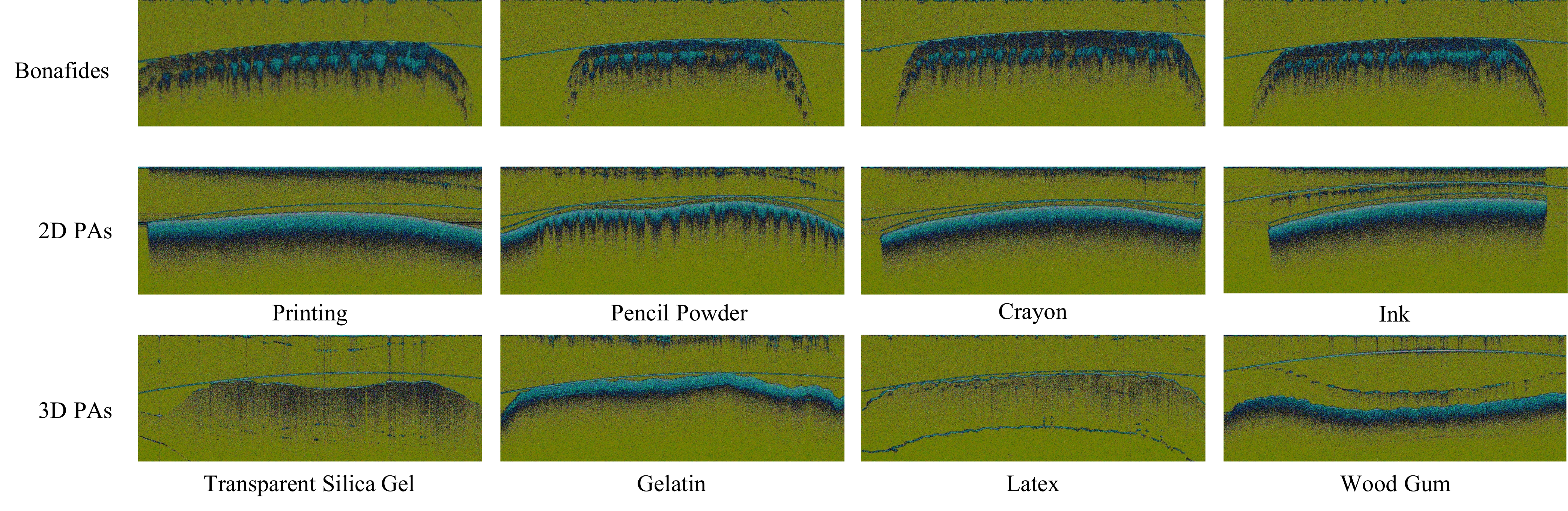}
  \caption{Some samples of bonafides and presentation attacks captured by OCT-based device. The first line represents the bonafide OCT-based fingerprints. The second line shows the 2D presentation attacks made from different materials. The third line gives the 3D presentation attacks. }
  \label{fig:real_fake}
\end{figure*}

To solve these restrictions, other systems and sensors are applied, including high-resolution fingerprints~\cite{zhao2009direct,liu2011novel}, multiview 3D fingerprints~\cite{liu20143d,liu2015study} and Optical Coherence Tomography (OCT) based sensor~\cite{liu2019high,liu2020flexible,liu2021one}.
Zhao \textit{et al.}~\cite{zhao2009direct} applied high-resolution sensor to obtain pores on the fingertip which are non-existent in PAs, and designed a method to achieve pore recognition. Liu \textit{et al.}~\cite{liu2015study} obtained curvature features from 3D fingerprint captured by binocular stereo vision sensor which can used for recognition and PAD. However, the performance of them are still have been restricted by their low tolerance to poor quality images (e.g., altered, worn-out, ultra-wet/dry fingers) and spoof attacks (e.g., artificial fingerprints, fingerprint tampering) due to the nature of surface imaging of fingerprints.

\begin{figure}[ht]
  \centering
  \includegraphics[width=0.48\textwidth]{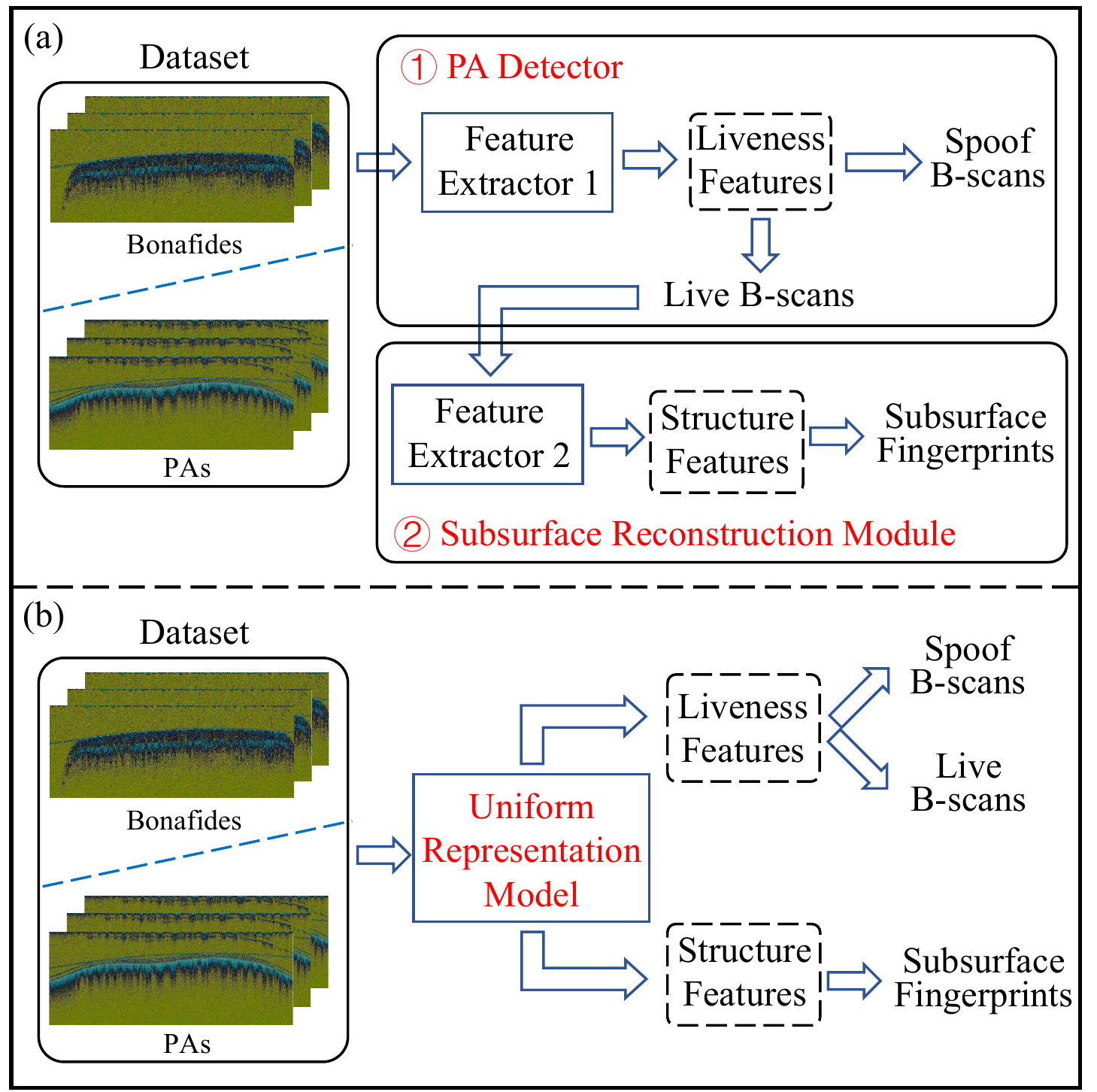}
  \caption{The pipelines to solve PAD and subsurface reconstruction problems of OCT-based fingerprint: (a).The process of existing methods. (b). The pipeline of our method using uniform representation model. }
  \label{fig:motivation}
\end{figure}

The introducing of technology of OCT~\cite{liu2020flexible} seems to provide a solution to solve the above problems since the information beneath the fingertip skin and the internal features of multilayered tissues can be captured.
As shown in Fig.~\ref{fig:oct}, OCT-based fingerprints provide subsurface structures of fingertips in the form of B-scans. Each B-scan, namely a cross-section slice, is the depth-wise (X-Z) representation of a fingertip. 
With OCT sensor scanning along the horizontal direction (X-Y) of fingertip, 400 B-scans are obtained to represent an instance of OCT-based fingerprint. A B-scan is formed by A-lines which image subsurface characteristics, and reach a penetration depth of the skin. As a bonafide B-scan example shown in Fig.~\ref{fig:oct} (c) and (d), OCT-based fingerprint contains depth information (i.e. multilayered skin tissues), including stratum corneum, viable epidermis, dermis and sweat duct. 
While in PA B-scans as shown in Fig.~\ref{fig:real_fake}, depth information is almost non-existent in their subsurface structures, no matter what type of presentation attack. 
Existing researches have demonstrated that these depth information of bonafide B-scans contains both liveness features (e.g. biological structures) and matching features (e.g. minutiae and pores)~\cite{liu2020flexible,ding2020surface}. 

For OCT-based fingerprint, how to utilize the depth information to achieve high accuracy PAD and obtain high quality reconstruction images are two key problems needed to solve. 
As shown in Fig.~\ref{fig:motivation}(a), existing researches on OCT-based fingerprint 
apply two independent modules and extract different features from depth information to solve the two key problems. They use a PA detector to obtain liveness features for PAD~\cite{liu2019high,liu2021one} and a reconstruction module to acquire the structure features for subsurface reconstruction~\cite{liu2020robust,ding2020surface}.  
However, the two independent steps will bring extra calculation and time cost and are inconsistent with real application scenarios. 
Using one model to extract liveness features and structure features at the same time is more expected in the application scenarios.
To tackle such problem, we attempt to establish a uniform representation model based on depth information as shown in Fig~\ref{fig:motivation}(b). Through only one model, the PAD can be completed and the obtained structure features can be further applied for subsurface reconstruction.

This paper proposes a uniform representation model of OCT-based fingerprint for PAD and reconstruction. Due to the liveness features and matching features both exist in the biological structures of OCT-based fingerprint, we design a novel semantic segmentation network to extract the multilayered skin tissues of B-scans. Through learning depth information of B-scans, a latent code in the segmentation model with abundant subsurface biological features is constructed. To achieve better generalization capability, the proposed model only requires bonafides for training. Thus, the latent code contains only liveness features which can be applied for PAD. Then, segmented B-scans are reconstructed to three 2D subsurface fingerprints with different depth information, which has the same patterns as surface fingerprints but with different image quality. Better performance is finally obtained when existing recognition techniques are directly adopted to those subsurface fingerprints.

The paper is organized as follows. In Section~\ref{sec:RW}, we review the state-of-the-art researches of OCT-based fingerprints. 
% and the mainstream semantic segmentation networks. 
Then, in Section~\ref{sec:method}, we introduce our proposed uniform representation model. Section~\ref{sec:exp} shows the evaluation of our proposed method and the comparison results with existing methods in PAD and recognition. Finally, our work is concluded in Section~\ref{sec:Con}.

\section{Related Work}
\label{sec:RW}
Since the key technologies related with our method are OCT-based fingerprint PAD and subsurface fingerprint reconstruction, in this section, we reviewed both of current researches and summarized in Table~\ref{tab:studies}. We then detailedly introduced them in the following Section ~\ref{sec:pad} and Section ~\ref{sec:feature}. 

\subsection{OCT-based Fingerprint PAD methods}
\label{sec:pad}
In general, there are two kinds of OCT-based fingerprint PAD methods, i.e. feature-based methods ~\cite{cheng2006artificial,darlow2016automated,liu2019high} and learning-based methods~\cite{nogueira2016fingerprint,chugh2019oct,liu2021one}. Feature-based approaches usually adopt hand-crafted features extracted by descriptor or histogram to distinguish bonafides and PAs. Darlow \textit{et al.}~\cite{darlow2016automated} extracted two anti-spoofing features through depth representation of A-lines, which are double-bright-feature and spoof-identification-feature, to differ bonafides and PAs. This method can achieve 100\% accuracy. However, this method is only evaluated on few PAs and applied in touchless OCT-based fingerprints. It will fail to detect PAs when using samples captured by touch-based OCT device.  
To solve this problem, Liu \textit{et al.}~\cite{liu2019high} proposed another two features, namely depth-double-peak and sub-single-peak, which can achieve PAD in touch-based OCT fingerprints. Through statistically analyzing the ratio of the two proposed features, bonafides and PAs can be confirmed. Their results are with 100\% accuracy when evaluating on four types of PA. 
However, feature-based methods are sensitive to image noise and may be failed if sample dataset beyond prior distribution resulting in the development of learning-based methods.

Compared with feature-based methods, learning-based methods can obtain more robust representation for PAD. Chugh \textit{et al.}~\cite{chugh2019oct} proposed to use a deep convolutional neural network (CNN) trained by bonafides and PAs for PAD. They extracted local patches of a B-scan as candidates and then input them to a CNN model for classification. Their method achieves a 1-Bonafide Presentation Classification Error Rate (BPCER) of 99.73\% @ Attack Presentation Classification Error Rate (APCER) of 0.2\% on a database of 3413 bonafide and 357 PA OCT scans. 
However, it is still lake of generalization to unkown PAs, since such kind of learning-based models heavily rely on the limited PAs in training set. 
Thus, Liu \textit{et al.}~\cite{liu2021one} designed a one-class PAD method, which trained only by bonafides, to tackle this problem. They proposed to represent bonafide distribution through an auto-encoder network and extracted reconstruction error and latent code to calculate the spoof score for PAD. Their method achieved 96.59\% 1-BPCER@APCER=5\% on a dataset with 93200 bonafide scans and 48400 PA scans.
However, we found that above learning-based methods did not consider the difference of depth information in different biological structure of subsurface fingerprints, which may improve the PAD performance. Thus, in this paper we attempt to distinguish the different subsurface layers by a semantic segmentation network and using its latent features to achieve PAD. 

\begin{table*}[ht]
\caption{Existing Studies on OCT-based Fingerprints}
\Huge
\centering
\resizebox{\textwidth}{!}{
\begin{tabular}{c|c|c|c|c|c|c}
\hline
\textbf{Task}                                                                      & \textbf{Study}                                                                                              & \textbf{Approach}                                                                                                                                                                                   & \textbf{OCT Technology}                                                                                    & \textbf{Database}                                                                                                                                       & \textbf{Category}                                                                                               & \textbf{Comments}                                                                                                                                                            \\ \hline
\multirow{22}{*}{PAD}                                                              & \multirow{6}{*}{\begin{tabular}[c]{@{}c@{}}Liu et\\ al. 2019\\ \cite{liu2019high}\end{tabular}}                & \multirow{6}{*}{\begin{tabular}[c]{@{}c@{}}Designed two anti-spoofing\\  features, namely depth \\ double peak feature \\ and sub-single-peak feature,\\  to represent bonafides.\end{tabular}}     & \multirow{6}{*}{\begin{tabular}[c]{@{}c@{}}Custom \\  Spectral-domain \\  OCT\end{tabular}}                & \multirow{6}{*}{\begin{tabular}[c]{@{}c@{}}Bonafide: 30 B-scans \\  from 15 subjects; \\ PA: 60 scans \\ from 4 PA materials.\end{tabular}}             & \multirow{6}{*}{\begin{tabular}[c]{@{}c@{}}Feature-based\\ method\end{tabular}}                                 & \multirow{6}{*}{\begin{tabular}[c]{@{}c@{}}Achieve 100\% \\ PAD accuracy.\end{tabular}}                                                                                      \\
                                                                                   &                                                                                                             &                                                                                                                                                                                                     &                                                                                                            &                                                                                                                                                         &                                                                                                                 &                                                                                                                                                                              \\
                                                                                   &                                                                                                             &                                                                                                                                                                                                     &                                                                                                            &                                                                                                                                                         &                                                                                                                 &                                                                                                                                                                              \\
                                                                                   &                                                                                                             &                                                                                                                                                                                                     &                                                                                                            &                                                                                                                                                         &                                                                                                                 &                                                                                                                                                                              \\
                                                                                   &                                                                                                             &                                                                                                                                                                                                     &                                                                                                            &                                                                                                                                                         &                                                                                                                 &                                                                                                                                                                              \\
                                                                                   &                                                                                                             &                                                                                                                                                                                                     &                                                                                                            &                                                                                                                                                         &                                                                                                                 &                                                                                                                                                                              \\ \cline{2-7} 
                                                                                   & \multirow{5}{*}{\begin{tabular}[c]{@{}c@{}}Chugh et\\ al. 2019\\ \cite{chugh2019oct}\end{tabular}}             & \multirow{5}{*}{\begin{tabular}[c]{@{}c@{}}Extracted local patches \\ of B-scan as input to\\  CNN model to distinguish\\  bonafides and PAs.\end{tabular}}                                         & \multirow{5}{*}{\begin{tabular}[c]{@{}c@{}}HORLabs\\  Spectral-domain\\  OCT (TEL1325LV2)\end{tabular}}    & \multirow{5}{*}{\begin{tabular}[c]{@{}c@{}}Bonafide: 3413 B-scans \\  from 415 subjects; \\  PA: 357 scans\\ from 8 PA materials.\end{tabular}}         & \multirow{5}{*}{\begin{tabular}[c]{@{}c@{}}Learning-based\\  method\end{tabular}}                               & \multirow{5}{*}{\begin{tabular}[c]{@{}c@{}}Use both PAs and \\  Bonafides for the \\  CNN training.\end{tabular}}                                                            \\
                                                                                   &                                                                                                             &                                                                                                                                                                                                     &                                                                                                            &                                                                                                                                                         &                                                                                                                 &                                                                                                                                                                              \\
                                                                                   &                                                                                                             &                                                                                                                                                                                                     &                                                                                                            &                                                                                                                                                         &                                                                                                                 &                                                                                                                                                                              \\
                                                                                   &                                                                                                             &                                                                                                                                                                                                     &                                                                                                            &                                                                                                                                                         &                                                                                                                 &                                                                                                                                                                              \\
                                                                                   &                                                                                                             &                                                                                                                                                                                                     &                                                                                                            &                                                                                                                                                         &                                                                                                                 &                                                                                                                                                                              \\ \cline{2-7} 
                                                                                   & \multirow{6}{*}{\begin{tabular}[c]{@{}c@{}}Liu et \\ al. 2021\\ \cite{liu2021one}\end{tabular}}                & \multirow{6}{*}{\begin{tabular}[c]{@{}c@{}}Designed a one-class \\ auto-encoder to represent\\  bonafide distribution. Extracted\\  reconstruction error and \\ latent code  for PAD.\end{tabular}} & \multirow{6}{*}{\begin{tabular}[c]{@{}c@{}}Custom \\  Spectral-domain \\  OCT\end{tabular}}                & \multirow{6}{*}{\begin{tabular}[c]{@{}c@{}}Bonafide: 93200 B-scans \\  from 176 subjects;\\  PA: 48400 PA scans\\ from 101 PA materials.\end{tabular}}  & \multirow{6}{*}{\begin{tabular}[c]{@{}c@{}}Learning-based\\  method\end{tabular}}                               & \multirow{6}{*}{\begin{tabular}[c]{@{}c@{}}Only use Bonafides\\ as the training set.\end{tabular}}                                                                           \\
                                                                                   &                                                                                                             &                                                                                                                                                                                                     &                                                                                                            &                                                                                                                                                         &                                                                                                                 &                                                                                                                                                                              \\
                                                                                   &                                                                                                             &                                                                                                                                                                                                     &                                                                                                            &                                                                                                                                                         &                                                                                                                 &                                                                                                                                                                              \\
                                                                                   &                                                                                                             &                                                                                                                                                                                                     &                                                                                                            &                                                                                                                                                         &                                                                                                                 &                                                                                                                                                                              \\
                                                                                   &                                                                                                             &                                                                                                                                                                                                     &                                                                                                            &                                                                                                                                                         &                                                                                                                 &                                                                                                                                                                              \\
                                                                                   &                                                                                                             &                                                                                                                                                                                                     &                                                                                                            &                                                                                                                                                         &                                                                                                                 &                                                                                                                                                                              \\ \cline{2-7} 
                                                                                   & \multirow{5}{*}{\begin{tabular}[c]{@{}c@{}}Zhang et\\ al. 2021\\ \cite{zhang2021fingerprint}\end{tabular}}     & \multirow{5}{*}{\begin{tabular}[c]{@{}c@{}}Designed a  CNN-based \\ model to obtain four\\  different frequency features\\  of  B-scans for PAD.\end{tabular}}                                      & \multirow{5}{*}{\begin{tabular}[c]{@{}c@{}}Custom \\  Spectral-domain \\  OCT\end{tabular}}                & \multirow{5}{*}{\begin{tabular}[c]{@{}c@{}}Bonafide: 93200 B-scans \\  from 176 subjects; \\ PA: 48400 PA scans\\ from 101 PA materials.\end{tabular}}  & \multirow{5}{*}{\begin{tabular}[c]{@{}c@{}}Learning-based\\  method\end{tabular}}                               & \multirow{5}{*}{\begin{tabular}[c]{@{}c@{}}Explore anti-spoofing \\  featurs in \\  frequency domain.\end{tabular}}                                                          \\
                                                                                   &                                                                                                             &                                                                                                                                                                                                     &                                                                                                            &                                                                                                                                                         &                                                                                                                 &                                                                                                                                                                              \\
                                                                                   &                                                                                                             &                                                                                                                                                                                                     &                                                                                                            &                                                                                                                                                         &                                                                                                                 &                                                                                                                                                                              \\
                                                                                   &                                                                                                             &                                                                                                                                                                                                     &                                                                                                            &                                                                                                                                                         &                                                                                                                 &                                                                                                                                                                              \\
                                                                                   &                                                                                                             &                                                                                                                                                                                                     &                                                                                                            &                                                                                                                                                         &                                                                                                                 &                                                                                                                                                                              \\ \hline
\multirow{22}{*}{Reconstruction}                                                   & \multirow{6}{*}{\begin{tabular}[c]{@{}c@{}}Darlow et\\ al. 2015\\ \cite{darlow2015efficient}\end{tabular}}     & \multirow{6}{*}{\begin{tabular}[c]{@{}c@{}}Used Sobel edge detection \\ method to extract the \\ contour of stratum corneum \\ and viable epidermis \\  junction for reconstruction.\end{tabular}}  & \multirow{6}{*}{\begin{tabular}[c]{@{}c@{}}THORLabs \\ Swept-source OCT\\  (OCS1300SS)\end{tabular}}       & \multirow{6}{*}{\begin{tabular}[c]{@{}c@{}}13 OCT volumes \\  from 10 fingers.\end{tabular}}                                                            & \multirow{6}{*}{\begin{tabular}[c]{@{}c@{}}Edge detection-\\  based method\end{tabular}}                        & \multirow{6}{*}{\begin{tabular}[c]{@{}c@{}}Sensitive to noise \\ and  image quality.\end{tabular}}                                                                           \\
                                                                                   &                                                                                                             &                                                                                                                                                                                                     &                                                                                                            &                                                                                                                                                         &                                                                                                                 &                                                                                                                                                                              \\
                                                                                   &                                                                                                             &                                                                                                                                                                                                     &                                                                                                            &                                                                                                                                                         &                                                                                                                 &                                                                                                                                                                              \\
                                                                                   &                                                                                                             &                                                                                                                                                                                                     &                                                                                                            &                                                                                                                                                         &                                                                                                                 &                                                                                                                                                                              \\
                                                                                   &                                                                                                             &                                                                                                                                                                                                     &                                                                                                            &                                                                                                                                                         &                                                                                                                 &                                                                                                                                                                              \\
                                                                                   &                                                                                                             &                                                                                                                                                                                                     &                                                                                                            &                                                                                                                                                         &                                                                                                                 &                                                                                                                                                                              \\ \cline{2-7} 
                                                                                   & \multirow{5}{*}{\begin{tabular}[c]{@{}c@{}}Sekulska et\\ al. 2017\\ \cite{sekulska2017detection}\end{tabular}} & \multirow{5}{*}{\begin{tabular}[c]{@{}c@{}}Extracted stratum corneum \\ and viable epidermis\\  junction by extracting two\\  peak signals of each A-line.\end{tabular}}                            & \multirow{5}{*}{\begin{tabular}[c]{@{}c@{}}Wasatch Photonics Inc \\  Spectral-domain \\  OCT\end{tabular}} & \multirow{5}{*}{\begin{tabular}[c]{@{}c@{}}4 OCT volumes \\  from 2 subjects.\end{tabular}}                                                             & \multirow{5}{*}{\begin{tabular}[c]{@{}c@{}}Peak detection-\\  based method\end{tabular}}                        & \multirow{5}{*}{\begin{tabular}[c]{@{}c@{}}Prove the same pattern \\  of surface fingerprint\\  and internal fingerprint.\end{tabular}}                                      \\
                                                                                   &                                                                                                             &                                                                                                                                                                                                     &                                                                                                            &                                                                                                                                                         &                                                                                                                 &                                                                                                                                                                              \\
                                                                                   &                                                                                                             &                                                                                                                                                                                                     &                                                                                                            &                                                                                                                                                         &                                                                                                                 &                                                                                                                                                                              \\
                                                                                   &                                                                                                             &                                                                                                                                                                                                     &                                                                                                            &                                                                                                                                                         &                                                                                                                 &                                                                                                                                                                              \\
                                                                                   &                                                                                                             &                                                                                                                                                                                                     &                                                                                                            &                                                                                                                                                         &                                                                                                                 &                                                                                                                                                                              \\ \cline{2-7} 
                                                                                   & \multirow{6}{*}{\begin{tabular}[c]{@{}c@{}}Liu et\\ al. 2020\\ \cite{liu2020robust}\end{tabular}}              & \multirow{6}{*}{\begin{tabular}[c]{@{}c@{}}Designed a projection\\  based reconstruction \\ method by accumulating\\  skin tissues between\\  different peak signals.\end{tabular}}                 & \multirow{6}{*}{\begin{tabular}[c]{@{}c@{}}Custom \\  Spectral-domain \\  OCT\end{tabular}}                & \multirow{6}{*}{\begin{tabular}[c]{@{}c@{}}300 OCT volumes \\  from 150 fingers.\end{tabular}}                                                          & \multirow{6}{*}{\begin{tabular}[c]{@{}c@{}}Peak detection-\\  based method\end{tabular}}                        & \multirow{6}{*}{\begin{tabular}[c]{@{}c@{}}Achieve reconstruction \\  and 8.05\% EER \\  of recognition.\end{tabular}}                                                       \\
                                                                                   &                                                                                                             &                                                                                                                                                                                                     &                                                                                                            &                                                                                                                                                         &                                                                                                                 &                                                                                                                                                                              \\
                                                                                   &                                                                                                             &                                                                                                                                                                                                     &                                                                                                            &                                                                                                                                                         &                                                                                                                 &                                                                                                                                                                              \\
                                                                                   &                                                                                                             &                                                                                                                                                                                                     &                                                                                                            &                                                                                                                                                         &                                                                                                                 &                                                                                                                                                                              \\
                                                                                   &                                                                                                             &                                                                                                                                                                                                     &                                                                                                            &                                                                                                                                                         &                                                                                                                 &                                                                                                                                                                              \\
                                                                                   &                                                                                                             &                                                                                                                                                                                                     &                                                                                                            &                                                                                                                                                         &                                                                                                                 &                                                                                                                                                                              \\ \cline{2-7} 
                                                                                   & \multirow{5}{*}{\begin{tabular}[c]{@{}c@{}}Ding et\\ al. 2021\\ \cite{ding2020surface}\end{tabular}}           & \multirow{5}{*}{\begin{tabular}[c]{@{}c@{}}Used a U-net combined\\  with BDC-LSTM \\ and HDC for OCT\\  B-scan segmentation.\end{tabular}}                                                          & \multirow{5}{*}{\begin{tabular}[c]{@{}c@{}}Custom \\  Spectral-domain \\  OCT\end{tabular}}                & \multirow{5}{*}{\begin{tabular}[c]{@{}c@{}}1572 OCT volumes \\  from 210 fingers.\end{tabular}}                                                         & \multirow{5}{*}{\begin{tabular}[c]{@{}c@{}}Edge detection-\\  based method\end{tabular}}                        & \multirow{5}{*}{\begin{tabular}[c]{@{}c@{}}Extract biological tissue\\  contours and sweat glands.\end{tabular}}                                                             \\
                                                                                   &                                                                                                             &                                                                                                                                                                                                     &                                                                                                            &                                                                                                                                                         &                                                                                                                 &                                                                                                                                                                              \\
                                                                                   &                                                                                                             &                                                                                                                                                                                                     &                                                                                                            &                                                                                                                                                         &                                                                                                                 &                                                                                                                                                                              \\
                                                                                   &                                                                                                             &                                                                                                                                                                                                     &                                                                                                            &                                                                                                                                                         &                                                                                                                 &                                                                                                                                                                              \\
                                                                                   &                                                                                                             &                                                                                                                                                                                                     &                                                                                                            &                                                                                                                                                         &                                                                                                                 &                                                                                                                                                                              \\ \hline
\multirow{8}{*}{\begin{tabular}[c]{@{}c@{}}PAD\\ \&\\ Reconstruction\end{tabular}} & \multirow{8}{*}{\begin{tabular}[c]{@{}c@{}}Proposed\\ Method\end{tabular}}                                  & \multirow{8}{*}{\begin{tabular}[c]{@{}c@{}}Designed a uniform \\ representation model for \\  B-scan segmentation, \\ PAD and reconstruction.\end{tabular}}                                         & \multirow{8}{*}{\begin{tabular}[c]{@{}c@{}}Custom \\  Spectral-domain \\  OCT\end{tabular}}                & \multirow{4}{*}{\begin{tabular}[c]{@{}c@{}}Bonafide: 70400 B-scans \\ from 176 subjects; \\ PAs: 48400 PA scans \\  from 101 PA matirelas\end{tabular}} & \multirow{8}{*}{\begin{tabular}[c]{@{}c@{}}Learning-based\\  method \&\\  Segmentation \\  method\end{tabular}} & \multirow{8}{*}{\begin{tabular}[c]{@{}c@{}}Achieve 96.63\% accuracy \\ of PAD, 2.25\% EER of \\  mitnutiae recognition \\  and 5.42\% EER of \\  pore recognition.\end{tabular}} \\
                                                                                   &                                                                                                             &                                                                                                                                                                                                     &                                                                                                            &                                                                                                                                                         &                                                                                                                 &                                                                                                                                                                              \\
                                                                                   &                                                                                                             &                                                                                                                                                                                                     &                                                                                                            &                                                                                                                                                         &                                                                                                                 &                                                                                                                                                                              \\
                                                                                   &                                                                                                             &                                                                                                                                                                                                     &                                                                                                            &                                                                                                                                                         &                                                                                                                 &                                                                                                                                                                              \\ \cline{5-5}
                                                                                   &                                                                                                             &                                                                                                                                                                                                     &                                                                                                            & \multirow{4}{*}{\begin{tabular}[c]{@{}c@{}}2136 OCT volumes \\  of 1068 fingers \\  from 136 subjects.\end{tabular}}                                    &                                                                                                                 &                                                                                                                                                                              \\
                                                                                   &                                                                                                             &                                                                                                                                                                                                     &                                                                                                            &                                                                                                                                                         &                                                                                                                 &                                                                                                                                                                              \\
                                                                                   &                                                                                                             &                                                                                                                                                                                                     &                                                                                                            &                                                                                                                                                         &                                                                                                                 &                                                                                                                                                                              \\
                                                                                   &                                                                                                             &                                                                                                                                                                                                     &                                                                                                            &                                                                                                                                                         &                                                                                                                 &                                                                                                                                                                              \\ \hline
\end{tabular}
}
\label{tab:studies}
\end{table*}

\begin{figure*}[t]
  \centering
  \includegraphics[width=0.98\textwidth]{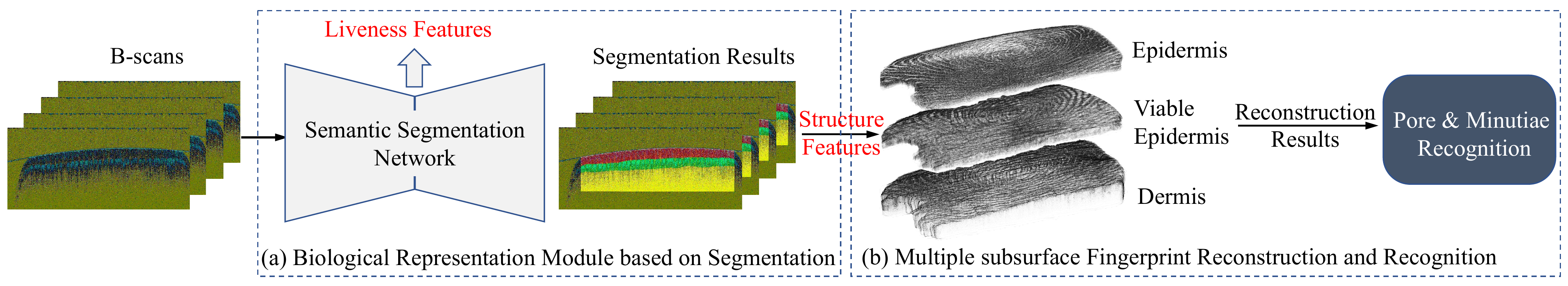}
  \caption{An overview of the proposed method for a whole fingertip. (a) B-scan representation based on Biological Segmentation. (b) Multiple subsurface Fingerprint Reconstruction and Recognition.}
  \label{fig:pipline}
\end{figure*}

\subsection{OCT-based Fingerprint Reconstruction Methods}
\label{sec:feature}

Since OCT-based fingerprint consists of a series cross-section slices, there are large differences of forms between the OCT-based fingerprints and traditional surface imaging fingerprints. To apply the existing recognition method (e.g. minutiae based recognition) for identification, it is necessary to reconstruct OCT-based fingerprint to 2D forms to achieve further recognition. Thus, the quality of reconstruction fingerprints is crucial for OCT-based recognition. 
At present, existing reconstruction methods can be divided into two categories, i.e. edge detection-based methods~\cite{khutlang2014novelty,darlow2015efficient,wang2018external,ding2020surface} and peak detection-based methods~\cite{sekulska2017detection,darlow2015internal,liu2020robust}.  
As shown in Table~\ref{tab:studies}, edge detection-based methods detect the contours of biological structure layers in each B-scan image. While, Peak detection-based methods detect skin tissues pixel by pixel for each biological layers in a B-scan image.
Among them, Liu \textit{et al.}~\cite{liu2020robust} designed a projection-based reconstruction method and proposed a fusion strategy to obtain the robust reconstructed fingerprints. However, the traditional edge detection-based methods are sensitive to the noise and image quality and peak detection-based methods can only detect the roughly range of different biological layers. 
To solve these problem, Ding \textit{et al.}~\cite{ding2020surface} proposed a modified U-net which combines residual learning, bidirectional convolutional long short-term memory and hybrid dilated convolution for B-scan segmentation and extract the contours of biological structure and sweat glands for reconstruction. 
However,~\cite{ding2020surface} is still limited by the edge detection-based method, which can only extract contours of biological layers, the depth information in the biological layers and whole tissue structures of B-scans were not considered. 

To tackle these issues, we considered whether a semantic segmentation network which can extract different biological regions will be an effective solution. There are many classical semantic segmentation networks. 
For example, Long \textit{et al.}~\cite{long2015fully} first designed a Fully Connected Network (FCN) to achieve semantic segmentation on public datasets. 
Ronneberger \textit{et al.}~\cite{ronneberger2015u} proposed an end-to-end network, namely U-net, which connected features between down sampling layers and up sampling layers to improve the segmentation performance on biomedical images.  
Chen \textit{et al.}~\cite{chen2018encoder} explored a DeepLabv3+ network, which combined atrous separable convolution and a decoder module for semantic image segmentation. 
Inspired by this, we attempt to design a semantic segmentation network adapting to the OCT-based fingerprint to acquire high quality subsurface biological regions.

As we discussed above, studies on OCT-based fingerprint PAD and reconstruction are completely independent. The extracted features cannot be represented in a common latent space for the two tasks. The depth information of OCT-based fingerprints is underutilized. Thus, considering the above problems, instead of detecting PAs and reconstructing B-scans in two separate stages, in this paper, we propose a uniform representation for OCT-based fingerprint to achieve PAD and reconstruction at the same time. 

As shown in Fig.~\ref{fig:pipline}, the proposed model consists of two module. The first module is a novel semantic segmentation network which apply attention mechanisms to attach biological structures to the segmentation results. 
The biological representation module is trained by bonafide B-scans and pixel-level annotations.
Through the trained segmentation network, the latent code and three biological layers in the B-scan subsurface can be extracted. 
Since the latent codes contain abundant live features and structure features which are non-existent in PAs, based on the latent code, we design an effective spoof score for PAD. Then, we straighten the segmented three-biological-layer B-scans to avoid the effect of skin stretching and accumulate them for reconstruction. Thus, we can obtain three reconstructed 2D subsurface fingerprints from stratum corneum, viable epidermis and dermis. Recognition are finally implemented by adopting existing classical minutiae-based and pore-based matchers to prove the superior quality of the reconstructed subsurface fingerprints compared with surface prints.

\begin{itemize}
  \item We for the first time propose a uniform representation model to achieve PAD and subsurface reconstruction for OCT-based fingerprint recognition systems. 
  \item A novel semantic segmentation network, which attaches the biological structure features to the segmentation result is designed using attention mechanisms.
  \item The superiority of our method are validated by comparing with the-state-of-art PAD methods and subsurface fingerprint reconstruction methods. Recognition performance evaluated on our reconstructed subsurface fingerprints, traditional 2D surface fingerprints and high resolution 2D surface ones further proved the effectiveness of the proposed method.
\end{itemize}

\begin{figure*}[ht]
  \centering
  \includegraphics[width=0.98\textwidth]{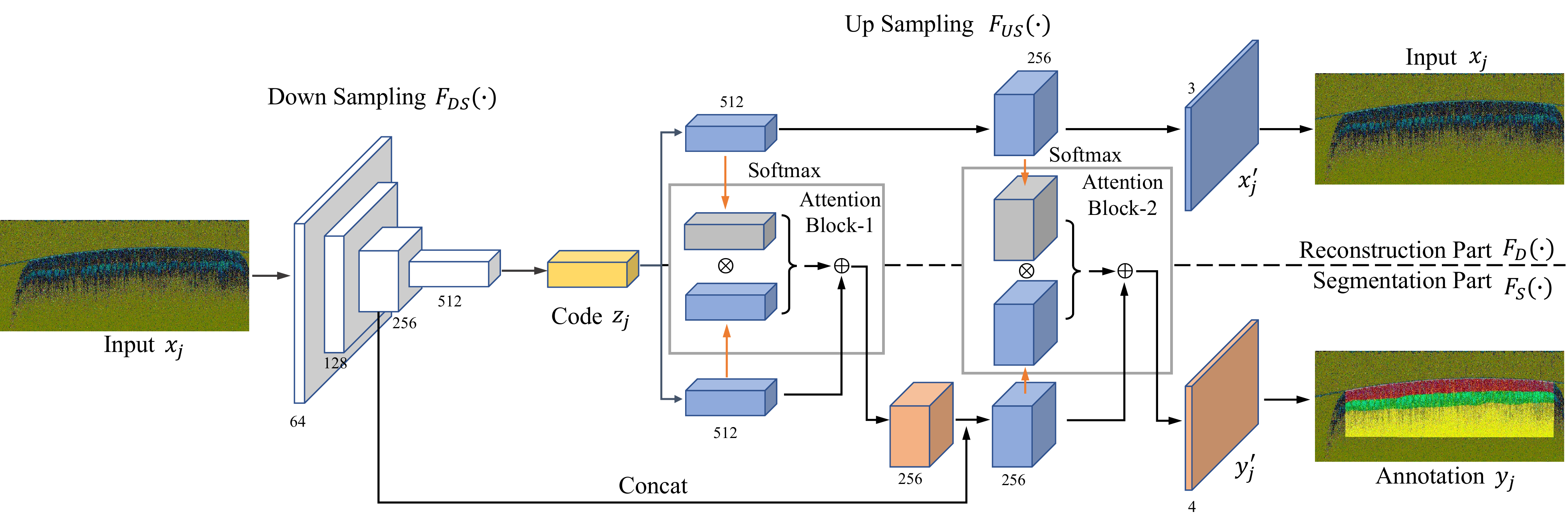}
  \caption{The network structure of the segmentation method.}
  \label{fig:network}
\end{figure*}

\section{Proposed Method}
\label{sec:method}
In this paper, we propose a uniform OCT-based fingerprint representation model (\textbf{U-OCFR}) by adopting depth information for PAD and subsurface fingerprint reconstruction. Fig.~\ref{fig:pipline} shows the flowchart of our proposed method. The biological representation module, which consists of segmentation network, is only trained by bonafide B-scans and their semantic annotations to learn the multilayered depth information. From the trained biological representation module, the latent codes after down sampling are applied to detect PAs. The multiple segmentation results are further applied to reconstruct subsurface fingerprints. The detailed implementation of the proposed method is introduced as follows.

\subsection{System Framework}
As shown in Fig.~\ref{fig:pipline}, our method mainly consists of two components, including a biological representation module, multilayered reconstruction and recognition. To better describe the proposed method, we first set up the basic notations in this paper. 
In Fig.~\ref{fig:pipline}(a), the biological representation module $S(\cdot)$ is first designed to obtain biological structures and extract latent codes for PAD at the same time. Through such module, PAs (spoof fingerprints) can be detected and three-layer biological structures of bonafides (live fingerprints) can be obtained. 
Then, in terms of the segmented biological layers, we design a reconstruction module $R(\cdot)$ to transform the 3D OCT-based fingerprints into three 2D fingerprints as shown in Fig.~\ref{fig:pipline}(b). By extracting pore and minutiae features in different reconstructed layers, the fingerprints are further matched. 

Specifically, given a set of training sample $S_{train}=\left \{ (X_i,Y_i)|i \in N_t \right \}$, the biological representation module $S(\cdot)$ can be trained using $N_t$ instances of OCT-based fingerprint $X_i$ and their corresponding biological structure annotations $Y_i$. For each instance $(X_i,Y_i)$, $\left \{(x_j,y_j) | j \in [1,400] \right \}$ represents B-scans (size of 500 $\times$ 1500 pixels) of one OCT-based fingerprint. 
Given an instance $X_i$ as the input of trained $S(\cdot)$, the latent codes $Z_i= \left \{z_1,...,z_j,...,z_{400} \right \}$ can be applied to detect PAs. 
When instance $X_i$ is detected as bonafide, the reconstruction module $R(\cdot)$ is designed to reconstruct OCT-based fingerprints to three different 2D fingerprints, i.e. stratum corneum, viable epidermis and dermis, respectively denoted as $L_S$, $L_V$, and $L_D$. These three fingerprints can be further applied for minutiae and pore recognition.

Due to the unique depth information existed in the subsurface biological structures of B-scans, it is crucial to learn and extract the biological representation of B-scans first.
Thus, the first step of our U-OCFR model is to train a B-scan representation based on biological segmentation as shown in Fig.~\ref{fig:pipline}(a). The output of biological representation module $S(\cdot)$ is the segmentation result in terms of the biological structures. The latent codes are as the high-level representation of $S(\cdot)$ to distinguish the disparities between PAs and bonafides.

\begin{table}[h]
\centering
\caption{Specifications for U-OCFR Segmentation Network}
\Huge
\resizebox{0.47\textwidth}{!}{
\begin{tabular}{c|c|c|c|c}
\hline
                               & \textit{Input}     & \textit{Output} & \textit{Operator}   & \textit{Block Num.} \\ \hline
\multirow{5}{*}{$F_{DS}(\cdot)$}      & 256$\times$768$\times$3  & 128$\times$384$\times$64    &\multirow{5}{*}{\begin{tabular}[c]{@{}c@{}}ResNet-block: \\ (conv2D (rate=1)+ \\ conv2D (rate=2)+ \\ conv2D (rate=5)+ \\ conv2D (strides=2))\end{tabular}}    & 1                \\
                               & 128$\times$384$\times$64 & 64$\times$192$\times$128    &                                 & 1                \\
                               & 64$\times$192$\times$128 & 32$\times$96$\times$256     &                                 & 1                \\
                               & 32$\times$96$\times$256  & 16$\times$48$\times$512     &                                 & 1                \\
                               & 16$\times$48$\times$512  & 8$\times$24$\times$512      &                                 & 1                \\ \hline
                               
\multirow{4}{*}{$Att(\cdot)$}         & 16$\times$48$\times$512  & \multirow{2}{*}{32$\times$96$\times$256}   &\multirow{4}{*}{\begin{tabular}[c]{@{}c@{}}Attention-block:\\ (Softmax+ \\ conv2D (strides=1)+ \\ Bilinear Resize)\end{tabular}} & \multirow{2}{*}{1}              \\
                               & 16$\times$48$\times$512  &                                            &         &                 \\\cline{2-3}\cline{5-5}
                               & 32$\times$96$\times$256  & \multirow{2}{*}{256$\times$768$\times$4}   &         & \multirow{2}{*}{1}         \\
                               & 32$\times$96$\times$256  &                                            &         &                 \\\hline

\multirow{3}{*}{$F_{D}(\cdot)$}       & 8$\times$24$\times$512  & 16$\times$48$\times$512   & Bilinear Resize            & 1               \\
                               & 16$\times$48$\times$512 & 32$\times$96$\times$256   & Bilinear Resize                               & 1               \\
                               & 32$\times$96$\times$256 & 256$\times$768$\times$3   & Bilinear Resize                      & 1               \\ \hline

\multirow{5}{*}{$F_{S}(\cdot)$}       & 8$\times$24$\times$512  & 16$\times$48$\times$512   & Bilinear Resize            & 1               \\
                               & 16$\times$48$\times$512 & 32$\times$96$\times$256   & Attention-block                               & 1               \\
                               & 32$\times$96$\times$256 & 32$\times$96$\times$256   & Concat                      & 1               \\
                               & 32$\times$96$\times$256 & 256$\times$768$\times$4   & Attention-block                               & 1               \\ \hline
\end{tabular}
}
\label{tab:structure}
\end{table}

\subsection{Biological Representation Module}
\subsubsection{Segmentation Network of U-OCFR model}
The network structure of our biological representation module is composed of Down Sampling $F_{DS}(\cdot)$ and Up Sampling $F_{US}(\cdot)$. 
$F_{DS}(\cdot)$ encodes the input B-scan $x_j$ and embed biological features into latent code $z_j$. 
Then, $F_{US}(\cdot)$ decodes these features to biological structure segmentation results $y_j$ and the original input $x_j$. Unlike other segmentation model, $F_{US}(\cdot)$ is composed of two parts, including Segmentation Part $F_{S}(\cdot)$ and Reconstruction Part $F_{D}(\cdot)$. In Segmentation Part $F_{S}(\cdot)$, in order to derive different perceptual fields of $x_j$, one of feature maps from $F_{DS}(\cdot)$ coupled with $z_j$ are regarded as the input of $F_{S}(\cdot)$. 
To obtain a more accurate result, we propose to apply an extra Reconstruction Part $F_{D}(\cdot)$ to strengthen the content and texture features learning. 
$F_{D}(\cdot)$ can reconstruct the latent code to input and the final segmentation result is predicted from the $F_{S}(\cdot)$. Two branches are connected by two Attention-blocks $Att(\cdot)$. Fig.~\ref{fig:network} shows the detail structures of the biological representation module. 

Specifically, the implementation of $S(\cdot)$ is listed in Table~\ref{tab:structure}. 
For $F_{DS}(\cdot)$, it consists of five ResNet-blocks and each block contains three atrous convolution layers and one convolution layer. 
For $F_{US}(\cdot)$, $F_{D}(\cdot)$ consits of three bilinear resize operation to upsamle the feature maps. 
$F_{S}(\cdot)$ consists of bilinear resize operation, two Attention-blocks and one layer to concat feature maps in $F_{DS}(\cdot)$. 
The trained $S(\cdot)$ can extract four semantic region of B-scans $y_j^{'}$, including  (1) stratum corneum region, (2) viable epidermis region, (3) dermis region and (4) background. 

The function of biological representation module is given by
\begin{equation}
S(x_j, y_j, \theta) =  F_{US}(F_{DS}(x_j))
  \label{eq:S}
\end{equation}
where the inputs of $S(\cdot)$ are B-scan $x_j$ and segmentation annotation $y_j$, $\theta$ refers to the learning parameters of network. The latent code $z_j$ can be obtained by
\begin{equation}
z_j = F_{DS}(x_j)
  \label{eq:DS}
\end{equation}
In Up Sampling part of $S(\cdot)$, the reconstruction result $x_{j}^{'}$ of input $x_j$ is given by
\begin{equation}
x_{j}^{'}  = F_D(z_j)
  \label{eq:de}
\end{equation}
where $x_{j}^{'}$ contains abundant content and texture features of B-scan biological structures.
Then, to attach the biological structure features to the segmentation result, we apply the Attention-blocks to connect $F_{D}(\cdot)$ and $F_{S}(\cdot)$. Each Attention-block is given by 
\begin{equation}
Att(f_{D_m}, f_{S_m})  = f_{S_m}  \otimes (1 + \frac{exp(f_{D_m})}{\sum_{c \in [1,n]}exp(f_{D_m}[c])})
  \label{eq:att}
\end{equation}
where $f_{D_m}$ and $f_{S_m}$ are the $m^{th}$ features in the middle of $F_{D}(\cdot)$ and $F_{S}(\cdot)$. 
We apply \textit{softmax} to normalize $f_{D_m}$ in channel dimension first. Then, multiply the normalized feature $f_{D_m}$ to $f_{S_m}$ and add it to $f_{S_m}$. The output of Attention-block is regarded as input of the next layer in $F_{S}(\cdot)$. 
Through applying Attention-blocks, the segmentation result $y_j^{'}=S(x_j, y_j, \theta)$ can be more accurate. 
In the training process, the reconstruction part from $x_{j}$ to $x_{j}^{'}$ is trained by a $L_2$ loss function
\begin{equation}
\mathcal{L}_{D} = \left \| x_j - x_{j}^{'} \right \|_2
  \label{eq:lD}
\end{equation}
The segmentation part from $y_{j}$ to $y_j^{'}$ is trained by a cross entropy as follows:
\begin{equation}
\mathcal{L}_{S} = - [y_jlog(y_j^{'})+(1-y_j)log(1-y_j^{'})]
  \label{eq:lS}
\end{equation}
where $y_j^{'}$ is the prediction result of $S(x_j, y_j, \theta)$. 
In sum, our biological representation module is trained by 
\begin{equation}
\mathcal{L} = \mathcal{L}_{D} + \mathcal{L}_{S}
  \label{eq:L}
\end{equation}

\subsubsection{Latent Code based Presentation Attack Detection}

Given $x_j$ as input to the trained biological representation module, latent code $z_j$ with size of $(8, 24, 512)$ can be obtained. However, such size of $z_j$ is too large, which will greatly increase computational cost when analyzing. Thus, we firstly use Global Average Pooling (GAP) in each feature map channel to resize $z_j$ to $1\times1\times512$, which denoted as $z_j^{'}$. 

Since our biological representation module is trained only by bonafides, the distribution of PAs is still unknown. 
To solve the problem, we set a bonafide reference set $S_{ref}=\left \{ X^r_i|i \in n \right \}$ different with $S_{train}$ to model the distribution of bonafides, like the strategy of our previous work~\cite{liu2021one}. 
The reference code $z_r$ is calculated by 
\begin{equation}
z_r = \frac{1}{400n}\sum_{x_j \in X^r_i}\sum_{j=1}^{400}F_{DS}(x_j)
  \label{eq:ref}
\end{equation}
In the same light, $z_r$ is averaged by GAP, which is denoted as $z_r^{'}$. 
Thus, latent code based PAD can be achieved through calculating disparities between $z_j^{'}$ and $z_r^{'}$ . 
On the basis of a instance of B-scans (a whole fingertip), We apply a spoof score to distinguish the differences:
\begin{equation}
Score_I = \frac{1}{400}\sum_{z_j \in Z_i} \left \| z_j^{'} - z_r^{'} \right \|_2
  \label{eq:instance}
\end{equation}
where $Score_I$ is calculated by the Euclidean distance from given latent code to reference code  when using instance with successive 400 B-scans of one finger as input. A large score represents the input is similar to presentation attack, otherwise, the input is close to a bonafide.

\subsection{Multiple Subsurface Fingerprint Reconstruction based on Uniform Representation}
Through the presentation attack detection, PAs can be segregated and only bonafide fingerprints are left for recognition. 
However, existing recognition algorithms are widely implemented in conventional 2D-form fingerprints. Fingerprints in B-scan forms cannot be matched directly. 
Thus, we firstly design a method to reconstruct fingerprint in B-scan forms to three 2D fingerprints based on the segmentation results $y_j^{'}$. Then such reconstructed fingerprints can be applied to match using existing recognition algorithms. 

Due to the segmented three-layer subsurface B-scans all containing features for recognition~\cite{liu2020flexible}, we design a biological semantic reconstruction method to leverage information of each layer. First, the Daubechies wavelet on level 2 is applied for the B-scans' denoising. 
Since the fingertip is pressed against the cover glass during the fingerprint collecting~\cite{liu2020flexible}, the biological skin tissues are stretched which will influence the quality of reconstructed fingerprints. Thus, it is crucial to straighten B-scans before reconstructing. Here, we use the projection method in ~\cite{liu2020robust} to recover the B-scans and their annotations, which are denoted as $x_{j}^{p}$ and $y_j^{'p}$. Among them, $m \times n$ represent the size of $x_{j}^{p}$, which are 256 $\times$ 768 pixels in this paper. For point $a$ in each straightened B-scan $x_{j(m \times n)}^{p}$, its corresponding annotation can be given by 
\begin{equation}
y_j^{'p}[h]= \left\{\begin{matrix}
1, & a \in h\\ 
0, & a \notin h
\end{matrix}\right.
  \label{eq:annotation}
\end{equation}
where $h \in \left \{s,v,d \right \}$, $y_j^{'p}[s]$ is denoted as stratum corneum region, $y_j^{'p}[v]$ represents viable epidermis region and $y_j^{'p}[d]$ is for dermis region. 
Then, the reconstructed fingerprints $R_{h}(j,n)$ are summarized as follows  
\begin{equation}
    R_h(j,n)= \sum_{m=1}^{256}x_{j(m \times n)}^{p}y_j^{'p}[h]
  \label{eq:reconstruct}
\end{equation}
where $j \in [1,400]$, $R_h$ is the reconstructed fingerprint within $h$ region. 
Thus, three reconstructed 2D fingerprints $R_s, R_v$ and $R_d$ can be obtained. 
Since pores and minutiae, which can be used for recognition, are detected in OCT-based fingerprints, we will extract them in $R_h$ and give matching results in the experiment to evaluate the reconstruction result.

\begin{table*}[t]
\centering
\tiny
\caption{Data Description And Partition}
\resizebox{0.96\textwidth}{!}{
\begin{tabular}{cc|cc|c}
\hline
\multicolumn{2}{c|}{Task}                                                                                          & \multicolumn{2}{c|}{Data Partition}                                                                                                                                                                                                                   & Description                                                                                                                                                                                \\ \hline
\multicolumn{2}{c|}{\multirow{5}{*}{Presentation Attack Detection}}   & \multicolumn{1}{c|}{Reference Data}                                                                                      & Test Data                                                                                                                  & \multirow{5}{*}{\begin{tabular}[c]{@{}c@{}}Reference data are not \\ used for training and \\ different with data \\ of segmentation task\end{tabular}}                                    \\ \cline{3-4}
                                                                                &  & \multicolumn{1}{c|}{\multirow{4}{*}{\begin{tabular}[c]{@{}c@{}}Bonafide: 16 instances\\ from 8 fingers\end{tabular}}}   & \multirow{4}{*}{\begin{tabular}[c]{@{}c@{}}Bonafide: 176 instances\\ PA: 121 instances \\ from 101 materials\end{tabular}} &                                                                                                                                                                                            \\
                                                                                   &       & \multicolumn{1}{c|}{}                                                                                                    &                                                                                                                            &                                                                                                                                                                                            \\
                                                                              &           & \multicolumn{1}{c|}{}                                                                                                    &                                                                                                                            &                                                  \\
                                                                         &                  & \multicolumn{1}{c|}{}                                                                                                    &                                                                                                                            &                          \\ \hline
\multicolumn{1}{c|}{\multirow{11}{*}{Reconstruction}} & \multirow{6}{*}{Segmentation}                                                                 & \multicolumn{1}{c|}{Training Data}                                                                                       & Test Data                                                                                                                  & \multirow{6}{*}{\begin{tabular}[c]{@{}c@{}}Bonafide: 16 instances \\ $\times$ 400 = 6400 \\ B-scans with Annotations. \\ Using 5-fold strategies \\ for training and testing\end{tabular}} \\ \cline{3-4}
                                                  \multicolumn{1}{c|}{}             &  & \multicolumn{1}{c|}{\multirow{5}{*}{\begin{tabular}[c]{@{}c@{}}Bonafide:  5120 B-scans\\ with Annotations\end{tabular}}} & \multirow{5}{*}{\begin{tabular}[c]{@{}c@{}}Bonafide:  1280 B-scans\\ with Annotations\end{tabular}}                        &                                                                                                                                                                                            \\
                                                    \multicolumn{1}{c|}{}              &      & \multicolumn{1}{c|}{}                                                                                                    &                                                                                                                            &                                                                                                                                                                                            \\
                                                  \multicolumn{1}{c|}{}             &      & \multicolumn{1}{c|}{}                                                                                                    &                                                                                                                            &                                                                                                                                                                                            \\
                                                   \multicolumn{1}{c|}{}        &      & \multicolumn{1}{c|}{}                                                                                                    &                                                                                                                            &                                                                                              \\
                                                  \multicolumn{1}{c|}{}         &      & \multicolumn{1}{c|}{}  &   &       \\ \cline{2-5}
\multicolumn{1}{c|}{}& \multirow{5}{*}{Recognition} & \multicolumn{1}{c|}{Genuine Pairs}                                                                                      & Impostor Pairs                                                                                                              & \multirow{5}{*}{\begin{tabular}[c]{@{}c@{}}Collected fingerprints from \\ 1068 fingers of  73 female \\ subjects and 63 male subjects.\\ Each finger is collected twice.\end{tabular}}     \\ \cline{3-4}
                                                  \multicolumn{1}{c|}{}        &      & \multicolumn{1}{c|}{\multirow{4}{*}{1068 pairs}}                                                                         & \multirow{4}{*}{\begin{tabular}[c]{@{}c@{}}1068  $\times$ 1067 $\times$ 2 \\ = 2279112 pairs\end{tabular}}                 &                                                                                                                                                                                            \\
                                                   \multicolumn{1}{c|}{}&          & \multicolumn{1}{c|}{}                                                                                                    &                                                                                                                            &                                                                                                                                                                                            \\
                                                 \multicolumn{1}{c|}{}    &          & \multicolumn{1}{c|}{}                                                                                                    &                                                                                                                            & 
                                                                         \\
                                                \multicolumn{1}{c|}{}     &          & \multicolumn{1}{c|}{}   &                                                                                                                            &\\ \hline
\end{tabular}
}
\label{tab:dataset}
\end{table*}

\begin{table*}[t]
\centering
\caption{PAD Performance of Compared Networks and Existing Methods}
\tiny
\resizebox{0.79\textwidth}{!}{
\begin{tabular}{c|c|ccc}
\hline
\multicolumn{2}{c|}{}                                        & Acc(\%)      & \textit{BPCER}$_{10}$(\%)   & \textit{BPCER}$_{20}$(\%)    \\ \hline
\multirow{3}{*}{Existing   Methods}  & Feature  based Method~\cite{liu2019high}& 87.21         & 15.34          & 46.02 \\
& One-Class GAN~\cite{engelsma2019generalizing}        & 94.30          & 7.95          & 79.55         \\
& OCPAD~\cite{liu2021one}            & 96.30 & 0.57 & 3.41 \\ \hline
\multirow{4}{*}{Compared   Networks} & U-net~\cite{ronneberger2015u}
                  & 96.63          & 1.70          & 2.84 \\
& FCN~\cite{long2015fully}                    & 94.61          & 3.41          & 14.77          \\
& DeepLabV3+~\cite{chen2018encoder}             & 87.54         & 26.14          & 36.93   \\   
& Baseline  & 86.87         & 21.02          & 42.61    \\\hline
\multicolumn{2}{c|}{Proposed Method}    & \textbf{96.63} & \textbf{0.57} & 3.41   \\ \hline
\end{tabular}
}
\label{tab:pad_result}
\end{table*}

\section{Experimental Results and Analysis}
\label{sec:exp}
The performance of our method is evaluated from two aspects, namely PAD and reconstruction. In Section~\ref{sec:pad_exp}, we assess the effectiveness of PAD in biological representation module. In Section~\ref{sec:rec_exp}, the multiple subsurface fingerprint reconstruction results are compared with existing methods through segmentation performance of the multilayered biological structures and recognition results using existing recognition algorithms.
% To evaluate the performance of the proposed method, in this section, several experiments were carried on our three established datasets as shown in Table~\ref{tab:dataset}. 
% Firstly, in Section~\ref{sec:pad_exp}, we evaluate the effectiveness of PAD in biological representation module. 
% Then, the multiple subsurface fingerprint reconstruction results are compared with existing methods through segmentation performance of the multilayered biological structures and recognition results using existing recognition algorithms in Section~\ref{sec:rec_exp}. 

\subsection{Validation of Presentation Attack Detection}
\label{sec:pad_exp}

\subsubsection{Dataset and Implementation Details}
\label{sec:dataset2}
As shown in Table~\ref{tab:dataset}, the dataset consists of reference data and test data in presentation attack detection task. Reference data $S_{ref}$ is used to model the distribution of bonafides and calculate the reference code $z_r$. It is composed by 16 bonafide instances from 8 fingers. For each finger, it is scanned in two different sessions. Test data consists of 121 presentation attacks (121 instances) from 101 materials and 176 bonafide instances from 137 subjects. The PA materials in dataset includes 2D PAs and 3D PAs, as shown in Fig.~\ref{fig:pa}. For 2D PAs, materials like printing, pencil powder, crayon and ink are applied to collect spoof fingerprints. 
While, the material of 3D PAs includes transparent silica gel, gelatin, latex, wood gum and so on. Specifically, 40 3D PAs and 81 2D PAs are included in our dataset. For bonafides in test data, 176 bonafide instances are from 88 fingers. Each finger is also scanned in two different sessions. 
To evaluate the performance of PAD methods, three metrics are used in this experiment, including classification accuracy (Acc), Bonafide presentation classification error rate ($BPCER$) when the highest attack presentation classification error rate ($APCER_{AP}$) = 10\% ($BPCER_{10}$) and $BPCER$ when $APCER_{AP}$= 5\% ($BPCER_{20}$).  The smaller $BPCER_{10}$ and $BPCER_{20}$, the better performance of the model can be achieved.

\subsubsection{Performance Evaluation}
\label{sec:dataset2_}
\begin{figure*}[t]
  \centering
  \includegraphics[width=0.97\textwidth]{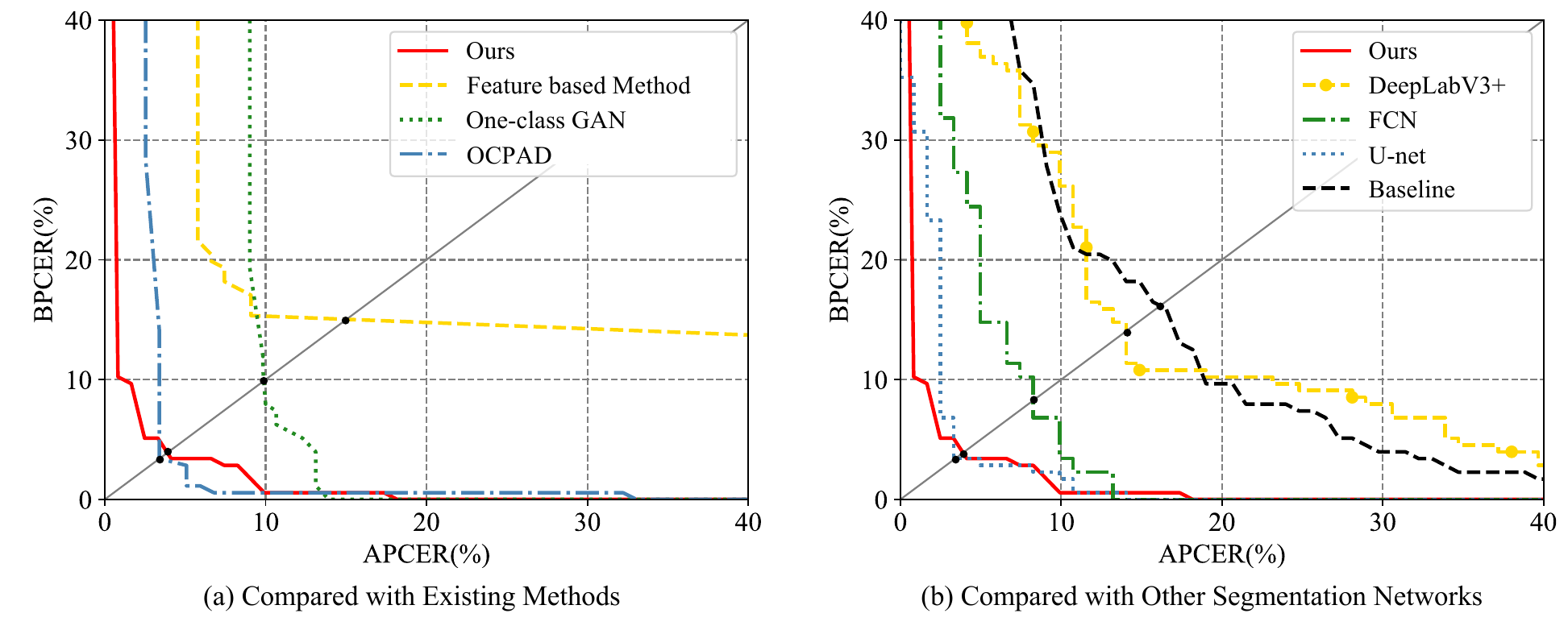}
  \caption{DET curves of PAD task on test data. (a) DET curve for our method and existing methods and (b) DET curve for our method and other Segmentation Networks.}
  \label{fig:det}
\end{figure*}

To validate the effectiveness of the proposed latent code based PAD method, a feature based PAD method~\cite{liu2019high}, a one-class GAN based method~\cite{engelsma2019generalizing} and the state-of-the-art method OCPAD~\cite{liu2021one} are included in this paper for comparison. Since segmentation networks for comparison in Section~\ref{sec:rec_exp} can extract latent codes for PAD, we also compared them by applying their latent codes, which has the same size with our method, to calculate the spoof score. 
As shown in Table~\ref{tab:pad_result}, the proposed method achieves 96.63\% Acc, which outperforms all the existing comparison methods. 
It is noted that one-class GAN and OCPAD methods are both learning based methods trained without using any PAs, which is the same with our method. However, their PAD performance is still lower than ours. 
The results indicate that more liveness features are aggregated into code $z_j$ through reconstruction and segmentation in the proposed biological representation module. The disparities between codes of bonafides and PAs are enlarged for more effective PAD. 
Meanwhile, compared with other segmentation networks, the proposed method achieves the best results with 96.63\% Acc and 0.57\% $BPCER_{10}$.
It shows that the PAD performance in the segmentation network benefits from the attention mechanisms of B-scan reconstruction part.
In our method, we used the latent codes generated from the semantic segmentation network for PAD. These high dimensional latent features have the merits of retaining more liveness features since they are generated from fine-grained semantic biological structures. However, existing PAD methods only rely on the coarse-grained histogram from depth signal or B-scan reconstruction, which ignored discriminative features for PAD in different subsurface layers.
Moreover, we draw the Detection Error Tradeoff (DET) curves to show more detailed information in Fig.~\ref{fig:det}. We can see that, the proposed method (red line in figure) outperforms all the comparison methods when APCER is small, which proves the superiority of our PAD method. When APCER is equal to BPCER, the Detection Equal Error Rate (D-EER) of our method can also reach a comparable result. Thus, for effective OCT-based PAD, it is important to extract features which retained more fine-gained semantic biological structures.

\subsection{Validation of Subsurface Fingerprint Reconstruction}
\label{sec:rec_exp}

\subsubsection{Dataset and Implementation Details}
\label{sec:dataset1}

\begin{table*}[h]
\centering
\tiny
\caption{Segmentation Performance of Compared Methods and Ablation Study}
\setlength{\arraycolsep}{1pt}
\resizebox{\textwidth}{!}{
\begin{tabular}{c|cc|cc|cc|cc|cc}
\hline
     & \multicolumn{2}{c|}{U-net~\cite{ronneberger2015u}} & \multicolumn{2}{c|}{FCN~\cite{long2015fully}}  & \multicolumn{2}{c|}{DeepLabV3+~\cite{chen2018encoder}} & \multicolumn{2}{c|}{Baseline} & \multicolumn{2}{c}{U-OCFR} \\ \hline
Num-Fold  & mIOU  & PA  & mIOU  & PA     & mIOU    & PA    & mIOU     & PA     & mIOU   & PA    \\ \hline
1    & 0.828 & 0.925   & 0.827 & 0.928    & 0.825   & 0.926     & 0.838    & 0.931    & \textbf{0.844}  & 0.933 \\
2    & 0.814 & 0.926  & 0.829 & 0.937     & 0.818   & 0.930     & 0.812    & 0.928    & 0.833  & 0.938 \\
3    & 0.814 & \underline{0.941}  & 0.802    & 0.936   & 0.803     & 0.939    & 0.832    & 0.948       & 0.830  & \textbf{0.945} \\
4    & 0.808 & 0.930   & 0.811 & 0.930    & 0.823   & 0.935     & 0.822    & 0.933   & 0.825  & 0.933 \\
5    & 0.821 & 0.929    & 0.838 & 0.935   &\underline{0.844}   & 0.939       & 0.828    & 0.933     & 0.840  & 0.936 \\ \hline
Mean & 0.817 & 0.930  & 0.821 & 0.933     & \underline{0.822}   & \underline{0.934}    & \textbf{0.826}    & \textbf{0.935}   & \textbf{0.834}  & \textbf{0.937} \\
Std. & 0.008 & 0.006   & 0.015 & 0.004   & 0.015   & 0.006     & 0.010    & 0.008     & \textbf{0.008}  & \textbf{0.005} \\ \hline
\end{tabular}
}
\label{tab:seg_result}
\end{table*}

As shown in Table~\ref{tab:dataset}, to validate the reconstruction performance of our method, we test segmentation results and recognition results based on reconstructed results in this section. 
In the segmentation task, we established a dataset containing 6400 bonafide B-scans from 16 instances with multilayered annotations (ground truth).
Each instance is collected from different fingers. which contains 400 B-scans. Each B-scan consists of 1500 A-lines and each A-line is composed by 500 pixels. Thus, a instance is with size of $400 \times 1500 \times 500$. To obtain the ground truth of multilayered structures, B-scans are annotated by three volunteers at the same time. Their average annotation results are used as the final ground truth to reduce the human error. 
To evaluate the segmentation network of biological representation module (U-OCFR), five-fold cross-validation strategy is applied in the experiment. The dataset is divided into five subsets. Four subsets are used for training and the last one are for testing. Each network is trained for 100 epochs. We use the best result for evaluation. Our U-OCFR model is trained by Adam with 1e-4 learning rate, 0.9 momentum and 5e-5 weight decay. Batch size for training is 16. We used two metrics, the mean Intersection of Union (mIOU) and Pixel Accuracy (PA) to verify the segmentation performance. Both metrics are within [0, 1]. The better segmentation performance performed, the higher mIOU and PA are obtained. This paper adopts Pytorch for all experiments using a work station with CPUs of 2.8GHz, RAM of 512GB and GPUs of NVIDIA Tesla V100.

In the recognition task, we adopted the dataset given in~\cite{liu2020flexible}. It contains 2136 fingerprints from 1068 fingers of 136 volunteers. Among them, the volunteers consist of 73 female subjects and 63 male subjects. Each finger was captured in two different sessions to form a match pair. Thus, there are 1068 genuine pairs and 2279112 impostor pairs. To evaluate the quality of our reconstructed subsurface fingerprints, we compared the recognition performance with 2D surface fingerprint. We collected the same fingers by two different devices, i.e. a U.are.U4000B commercial optical fingerprint sensor (to obtain 512 dpi, 328 $\times$ 356 pixels images) and a high-resolution  fingerprint sensor~\cite{zhao2009direct} (to obtain 1200 dpi, 640 $\times$ 480 pixels images). To keep the fairness of the experiment, the number of fingerprint captured by these two devices are also the same with OCT-based fingerprints. Two metrics are used in this experiment, including Equal Error Rate (EER) and FMR100~\cite{maio2002fvc2002}. EER is the value when false match rate is equal to false non-match rate. FMR100 measures the false non-match rate when false match rate is equal to 1/100. The smaller the EER or FMR100 is, the better the recognition performance is.
Besides, we apply Genuine Match Rate (GMR) when False Match Rate (FMR) = 5\% (GMR@FMR=5\%) and GMR@FMR=10\% metrics to evaluate the performance of feature extraction. The higher the value of GMR@FMR=5\% and GMR@FMR=10\% are, the better the performance is.
It is noted that, dataset in~\cite{liu2020flexible} contains a few fingerprints whose the pattern was damaged by some distortion, such as cuts, scars, and wear). The recognition metrics calculated by existing matching methods will be thus lower, since the number of genuine pairs in dataset~\cite{liu2020flexible} is still limited. 
\begin{figure*}[t]
  \centering
  \includegraphics[width=0.99\textwidth]{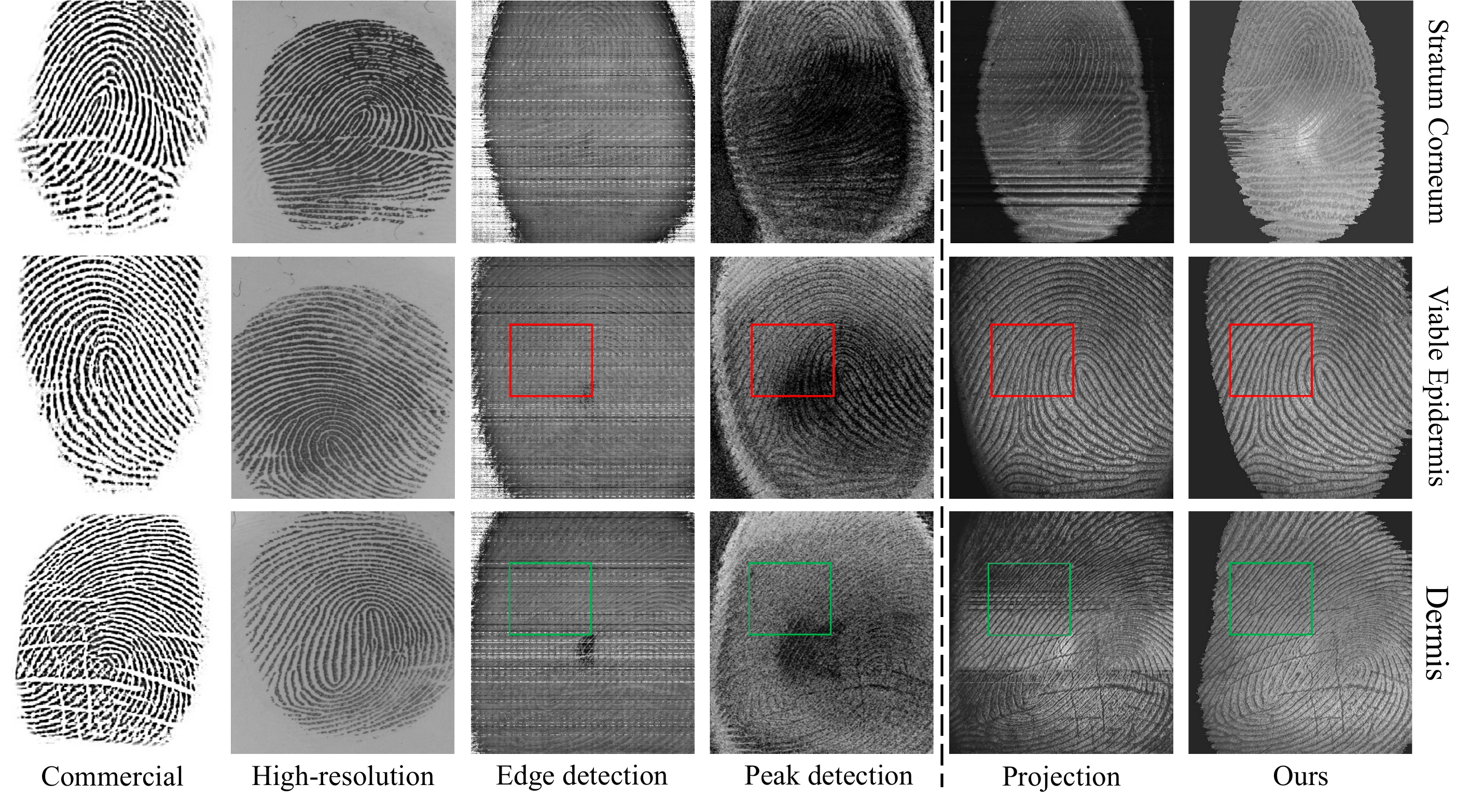}
  \caption{The reconstructed results using different methods and corresponding 2D surface fingerprints. We adopt three fingerprints from different fingers in each line. The images from left to right correspond to 2D surface fingerprints captured by commercial sensor and high resolution sensor; The reconstructed results using edge detection-based method, peak detection-based method, projection based method and ours. For multilayered reconstruction methods (projection based method and ours), the images from first line to third line correspond to stratum corneum, viable epidermis and dermis.}
  \label{fig:recon}
\end{figure*}

\subsubsection{Segmentation Performance Evaluation}
\label{sec:dataset1_}
To test the performance of the proposed segmentation network, three state-of-the-art networks, namely FCN~\cite{long2015fully}, U-net~\cite{ronneberger2015u}, and DeepLabV3+~\cite{chen2018encoder}, are included for comparison in this paper. 
To quantify the contribution of reconstruction part $F_D(\cdot)$ to the proposed U-OCFR model, the ablation study is also performed to evaluate performance of U-OCFR with or without $F_D(\cdot)$. To keep the fairness of the experiments, all the comparison methods and ablation study use the same protocol and training strategies with U-OCFR model.

\begin{table*}[t]
\centering
\caption{Minutiae Recognition Results Using different Reconstruction Methods and Devices
}
\tiny
\resizebox{0.75\textwidth}{!}{
\begin{tabular}{c|c|c|cc}
\hline
\multicolumn{3}{c|}{}                                                                 & EER(\%) & FMR100(\%) \\ \hline
\multirow{2}{*}{Other Devices}           & \multicolumn{2}{c|}{Commercial}             & 2.94     & \underline{4.54}       \\ \cline{2-5} 
                                         & \multicolumn{2}{c|}{High-resolution}        & \underline{2.63}    & 5.49       \\ \hline
\multirow{8}{*}{\begin{tabular}[c]{@{}c@{}}Reconstruction \\ Methods \end{tabular}} & \multicolumn{2}{c|}{Edge Detection~\cite{darlow2015efficient}}         & 47.36   & 96.80      \\ \cline{2-5} 
                                         & \multicolumn{2}{c|}{Peak Detection~\cite{sekulska2017detection}}         & 20.93   & 60.59      \\ \cline{2-5} 
                                         & \multirow{3}{*}{Projection~\cite{liu2020robust}} & Stratum Corneum     & 26.23    & 86.84       \\
                                         &                             & Viable Epidermis & 6.28    & 22.27       \\
                                         &                             & Dermis        & 10.40    & 86.02       \\ \cline{2-5} 
                                         & \multirow{3}{*}{Ours}     & Stratum Corneum     & 20.81   & 63.65      \\
                                         &                             & Viable Epidermis & \textbf{2.25}    & \textbf{4.29}       \\
                                         &                             & Dermis        & 6.31    & 13.35      \\ \hline
\end{tabular}
}
\label{tab:minutiae}
\end{table*}

\begin{table*}[h]
\centering
\caption{Pore Extraction Results Using different Reconstruction Methods and Devices}
\tiny
\resizebox{0.85\textwidth}{!}{
\begin{tabular}{c|c|c|cc}
\hline
\multicolumn{3}{c|}{}                                                                 & GMR@FMR=5\% & GMR@FMR=10\% \\ \hline
Other Devices                           & \multicolumn{2}{c|}{High-resolution}        & \underline{83.22}   & \underline{86.89}  \\ \hline
\multirow{6}{*}{\begin{tabular}[c]{@{}c@{}}Reconstruction \\ Methods \end{tabular}} 
                                        & \multirow{3}{*}{Projection~\cite{liu2020robust}} & Stratum Corneum     & 71.07   & 72.65  \\
                                        &                             & Viable Epidermis & 70.61   & 72.30  \\
                                        &                             & Dermis        & 68.22   & 71.60  \\ \cline{2-5} 
                                        & \multirow{3}{*}{Ours}     & Stratum Corneum     & \textbf{88.26}   & \textbf{89.66}  \\
                                        &                             & Viable Epidermis & 83.25   & 86.51  \\
                                        &                             & Dermis        & 82.70   & 84.63  \\ \hline
\end{tabular}
}
\label{tab:pore_extract}
\end{table*}

\begin{table*}[h]
\centering
\caption{Pore Recognition Results Using different Reconstruction Methods and Devices}
\tiny
\resizebox{0.8\textwidth}{!}{
\begin{tabular}{c|c|c|cc}
\hline
\multicolumn{3}{c|}{}                                                                 & \begin{tabular}[c]{@{}c@{}}In Layer \\ (EER(\%))\end{tabular} & \begin{tabular}[c]{@{}c@{}}Out of layer \\(EER(\%))\end{tabular} \\ \hline
Other Devices                            & \multicolumn{2}{c|}{High-resolution}        & \multicolumn{1}{c|}{\underline{5.55}}                                                  & -                               \\ \hline
\multirow{6}{*}{\begin{tabular}[c]{@{}c@{}}Reconstruction \\ Methods \end{tabular}}  
                                         & \multirow{3}{*}{Projection~\cite{liu2020robust}} & Stratum Corneum     & \multicolumn{1}{c|}{15.40}                                                 & 15.40                           \\
                                         &                             & Viable Epidermis & \multicolumn{1}{c|}{10.12}                                                 & 8.68                            \\
                                         &                             & Dermis        & \multicolumn{1}{c|}{12.93}                                                 & 12.4                            \\ \cline{2-5} 
                                         & \multirow{3}{*}{Ours}     & Stratum Corneum     & \multicolumn{1}{c|}{12.36}                                                 & 12.36                           \\
                                         &                             & Viable Epidermis & \multicolumn{1}{c|}{5.76}                                                  & \textbf{5.42}                            \\
                                         &                             & Dermis        & \multicolumn{1}{c|}{6.24}                                                  & 5.73                            \\ \hline
\end{tabular}
}
\label{tab:pore_reco}
\end{table*}
The five-fold segmentation results of comparison methods and ablation study of U-OCFR are shown in Table~\ref{tab:seg_result}. We calculate the mean and standard deviation (Std.) to evaluate the performance of five-fold experiments. Especially, the state-of-the-art method DeepLabV3+ can achieve 0.822 mIOU and 0.934 PA. While, U-OCFR can outperform DeepLabV3+ by 0.834 mIOU and 0.937 PA and achieve the best results. In the ablation study, we compared U-OCFR with Baseline. The Baseline represents U-OCFR model without using reconstruction part. As shown in Table~\ref{tab:seg_result}, the mean results of Baseline outperforms all comparison methods. By using reconstruction part, U-OCFR further improves the segmentation performance. It indicated that reconstruction part, which extracts the global biological structures from inputs, can be applied in the subsurface segmentation task. Through attaching reconstruction part as attention mechanisms to segmentation network, more accurate semantic regions are segmented. 
In addition, the Std. results of U-OCFR are lower than other methods, which shows our method achieves more stable performance in the five-fold experiments. It also proves the superiority of our method.

\subsubsection{Reconstruction Performance Evaluation}
\label{sec:dataset2_}
To validate the effectiveness of our method, existing reconstruction methods are compared in this experiment, including edge detection-based method~\cite{darlow2015efficient}, peak detection-based method~\cite{sekulska2017detection} and projection based method~\cite{liu2020robust}. The reconstructed fingerprints using different methods and fingerprint captured from other devices are shown in Fig.~\ref{fig:recon}. 
From the first line of Fig.~\ref{fig:recon}, the pore features can only be acquired from high-resolution fingerprint, projection-based method and our method. Meanwhile, compared with other surface fingerprints and reconstruction methods, only our method can both access the best image quality and fix the wrinkles and scratches of the surface fingerprint through applying the depth information of subsurface stratum corneum structures. In the second line, compared with other reconstruction methods, our method obtains a fingerprint with better contrast of ridge and valley lines. This is because more accurate biological structures of viable epidermis are segmented by our segmentation network. For instance in red boxes, less noises are generated by our method and clearer ridge and valley lines are shown. Similarly, in the third line, our method can also better reconstruct dermis layer, however, other methods cannot obtain clear images. In the green boxes, our method also obtains the best image quality. 
Therefore, It indicates the effectiveness and superiority of our reconstruction method. 

\subsubsection{Recognition Performance Evaluation based on the reconstructed subsurface fingerprint }
\label{sec:dataset3_}
Two recognition methods based on different level features are applied to match fingers. In detail, we apply NBIS software~\cite{maddala2011implementation} to extract minutiae features and match them. Then, we apply pore extraction method~\cite{zhao2010adaptive} to obtain pore features and match them using the method in~\cite{zhao2009direct}. 
The minutiae recognition results using different reconstruction methods and devices are shown in Table~\ref{tab:minutiae}. In viable epidermis layer, our method obtains 2.25\% EER and 4.29\% FMR100. 
Compared with recognition results using surface imaging fingerprints (other devices) and other reconstructed fingerprints, the proposed method obtains lower results in all three reconstructed subsurface layers, which achieves the best performance. Thus, It further demonstrate the effectiveness of our reconstruction method using subsurface biological structures.

As shown in Fig.~\ref{fig:oct}, since sweat duct can be captured in all subsurface layers in B-scans, pore recognition can be applied in the reconstructed OCT-based fingerprints. As shown in Fig.~\ref{fig:recon}, pores are clearly shown in the reconstructed stratum corneum layer (white points in ridge lines) and in another two layers, pores can also be captured (black points in ridge lines). Thus, we perform the pore extraction and recognition to prove the effectiveness of our reconstruction method. As shown in Table~\ref{tab:pore_extract}, pores can be extracted in all three subsurface layers. In high resolution fingerprints, the extraction results can achieve 83.22\% GMR@FMR=5\%. While, our method obtains the best extraction results with 88.26\% GMR@FMR=5\% and 89.66\% GMR@FMR=10\% in stratum corneum layer. 
Since the pore recognition method~\cite{zhao2009direct} will be invalid when the image quality is too low. As shown in Fig.~\ref{fig:recon}, ridge-valley lines in viable epidermis and dermis are more clear than in stratum corneum layer. To make full use of extracted pores, we set two experiments, namely in layer and out of layer. 
In layer experiments match fingers using pore extraction results in the corresponding subsurface layers. While, out of layer are only using pore extraction results in stratum corneum, and match them in each subsurface layers. 
As shown in Table~\ref{tab:pore_reco}, our method can achieve the best results with 5.42\% EER when using pore extraction results in stratum corneum and matching them in viable epidermis. This justifies that our reconstruction method can obtain higher quality fingerprint images.

\section{Conclusion}
\label{sec:Con}

To develop an OCT-based fingerprint system with high security, previous methods propose to detect PAs and reconstruct fingerprints in two separate stages. However, this will cost extra computational time in the real application scenarios.
Thus, this paper proposed a uniform representation model for OCT-based fingerprint to achieve PAD and reconstruction at the same time. We first design a novel semantic segmentation network by applying attention mechanisms from biological structures to extract the three biological layers in the B-scan subsurface. We then apply the latent codes from the trained segmentation network to achieve one-class presentation attack detection. Finally, the multiple segmentation results with biological structures are reconstructed to 2D fingerprints for the following recognition in an AFRS. 
In the experiment, PAD performance is evaluated by comparing the results of existing PAD methods and other segmentation networks. Our uniform representation model can obtain 96.63\% Acc and 0.57\% $BPCER_{10}$ which reach the state-of-the-art PAD performance. To evaluate the reconstruction performance, the segmentation, reconstruction, and recognition results are considered into experiments.
We compared results using other reconstruction methods and surface imaging fingerprints to validate the effectiveness of our method. In segmentation experiments, our method achieved the best segmentation results with 0.834 mIOU and 0.937 PA. In recognition experiments, the lowest results with 2.25\% EER of minutiae recognition and 5.42\% EER of pore recognition are achieved, which indicates the excellent reconstruction capability of the proposed uniform representation model. 

\section{Acknowledgements}

This work is supported in part by National Natural Science Foundation of China under Grant 62076163 and Grant 91959108; In part by the Shenzhen Fundamental Research Fund under Grant JCYJ20190808163401646; And in part by the Research Council of Norway (No. 321619 Project "OffPAD"). 

\bibliographystyle{IEEEtran}
\bibliography{myreference}
\end{document}